\documentclass[journal]{IEEEtran}
\usepackage{graphicx}
\usepackage{epstopdf}
\usepackage{subfigure}
\usepackage{amsmath,amssymb}
\usepackage{caption}
\DeclareCaptionLabelSeparator{periodquad}{.\quad}
\ifCLASSINFOpdf
\else
\fi
\begin{document}
%
\title{Few-Data Guided Learning Upon End-to-End Point Cloud Network for 3D Face Recognition}
%
%
%

\author{Yi Yu,
        Feipeng Da,
        and Ziyu Zhang
\thanks{This work was supported by Shenzhen Science and Technology Innovation Commission (STIC) under Grant JCYJ20180306174455080.}
\thanks{Yi Yu, Feipeng Da, and Ziyu Zhang are with the School of Automation, Southeast University, Nanjing 210096, China, and also with the Key Laboratory of Measurement and Control of Complex Systems of Engineering, Ministry of Education, Southeast University, Nanjing 210096, China (e-mail: dafp@seu.edu.cn).}
\thanks{}}

%
%

\markboth{}%
{}
%



\maketitle

\begin{abstract}
3D face recognition has shown its potential in many application scenarios. Among numerous 3D face recognition methods, deep-learning-based methods have developed vigorously in recent years. In this paper, an end-to-end deep learning network entitled Sur3dNet-Face for point-cloud-based 3D face recognition is proposed. The network uses PointNet as the backbone, which is a successful point cloud classification solution but does not work properly in face recognition. Supplemented with modifications in network architecture and a few-data guided learning framework based on Gaussian process morphable model, the backbone is successfully modified for 3D face recognition. Different from existing methods training with a large amount of data in multiple datasets, our method uses Spring2003 subset of FRGC v2.0 for training which contains only 943 facial scans, and the network is well trained with the guidance of such a small amount of real data. Without fine-tuning on the test set, the Rank-1 Recognition Rate (RR1) is achieved as follows: 98.85\% on FRGC v2.0 dataset and 99.33\% on Bosphorus dataset, which proves the effectiveness and the potentiality of our method.
\end{abstract}


%
\IEEEpeerreviewmaketitle

\section{Introduction}
%
%
%
%


\IEEEPARstart{W}{ith} the development of deep learning and 3D measurement technology, 3D face recognition has shown its potential in many application scenarios. Several solutions to 3D face recognition have been proposed including feature-based, model-based, matching-based, and learning-based methods, among which learning-based methods are flourishing in recent years and have shown remarkable performance \cite{Patil20153D, Soltanpour2017Survey}.

However, whether deep learning can achieve good results, to a great extent, depends on training data. As is known, many articles use more and more data to train the network. Although these methods compare the recognition rate on the same dataset, they do not use the same training set, which makes the comparison unfair. As a result, we can only see that the reported recognition rate improves again and again, but cannot distinguish whether it is the effect of the method itself or caused by the increase of training data. Take the state-of-the-art method proposed in \cite{Cai2019Fast} as an example, six datasets are used in the training process including about $22K$ scans, and the real data used for training are ten times more than those for testing, which makes it easy to obtain satisfactory recognition rate. However, in large-scale practical applications, the training set is usually smaller than the test set, and thus the recognition rate could be far below expectation, revealing that 3D face recognition is still an unsolved problem.

In this work, an end-to-end deep learning network entitled Sur3dNet-Face for point-cloud-based 3D face recognition is proposed. Taking advantage of a novel training framework upon Gaussian process morphable models (GPMM), and supplemented with a small amount of real data, the network is well trained.

The main contributions of this paper are as follows:

1) An end-to-end deep learning network for 3D face recognition is proposed to get face representations directly from 3D point clouds.

2) A few-data guided learning framework based on GPMM is established, upon that our network can be well trained with a small amount of real data.

3) The ablation study is analyzed to determine the parameters of the proposed network and the comparisons with other methods are conducted to prove the effectiveness and the potentiality of our method.

\section{Related Work}

\subsection{3D Face Recognition}

Numerous 3D face recognition methods have been developed, as reviewed in \cite{Patil20153D, Soltanpour2017Survey, Sun2014Deep}, including feature-based, model-based, matching-based, and learning-based methods.

The first three categories were widely studied in the early years. For example, Mian et al. \cite{Mian2008Keypoint} proposed a multimodal face recognition system, where the 3D and 2D faces are matched through modified ICP and SIFT descriptors respectively. Mohammadzade and Hatzinakos proposed Iterative Closest Normal Point (ICNP) \cite{Mohammadzade2013Iterative} to match face surfaces. Liu et al. \cite{Liu2013Learning} developed the harmonic feature-based approach using energies in spherical harmonics at different frequencies. Elaiwat et al. \cite{Elaiwat2015Curvelet} calculated Curvelet transform to extract features from semi-rigid regions. Lei et al. \cite{Lei2016Two} proposed a keypoint-based Multiple Triangle Statistics (KMTS) method to handle pose variations. 

As a latecomer, learning-based methods have gradually raised its head nowadays. Lei et al. \cite{Lei2014Efficient} trained the Kernel Principal Component Analysis (KPCA) to extract feature representations. Song et al. \cite{Song2018Dictionary} built 3D models to generate 2D images to improve the accuracy of 2D methods. Gilani and Mian \cite{Gilani2018Learning} and Kim et al. \cite{Kim2017Deep} employed existing 2D deep neural networks to solve the 3D face recognition problem by projecting 3D surface into 2D space as depth map, azimuth map, and elevation map. Cai et al. \cite{Cai2019Fast} proposed a deep learning technique based on facial component patches using depth map of the 3D face as the input of the traditional 2D network.

Some of these methods achieve decent accuracy, but the majority are still 2D-based networks, where 3D data are firstly projected into 2D images and then the traditional 2D networks are utilized to solve the problem.

\subsection{Point Cloud Network}

There are many representations of 3D data, among which point clouds are widely used. Different from traditional deep learning networks upon 2D images, Charles et al. proposed PointNet \cite{Charles2017PointNet} to directly handle point clouds, and the enhanced version PointNet++ \cite{Charles2017PointNetpp} is established upon PointNet along with grouping and sampling techniques to synthesize both global and local features of point clouds. 

Similarly, Li et al. proposed PointCNN \cite{Charles2017PointNet} to learn $\mathcal{X}$-transform so that the convolution operator can work on unordered point clouds, and Komarichev et al. proposed A-CNN computing convolution directly on point clouds through annular convolution.

Graph Neural Network (GNN) is also introduced into point cloud networks. For example, DGCNN \cite{Wang2019Dynamic} uses EdgeConv, a graph-based operation, to carry out convolution in feature space, and DPAM \cite{Liu2019Dynamic} uses a graph network to take the place of sampling and grouping step in PointNet++.

These methods are designed for object classiﬁcation and segmentation, though can also be applied to face recognition, the performance is much lower than face recognition methods mentioned in the previous subsection.

\subsection{Facial Data Generation}

Whether deep learning can achieve good results, to a great extent, depends on training data. Different from 2D face datasets, the 3D face data are relatively inadequate for training a network. 

Numerous 3D face generation methods have been developed. For example, Blanz and Vetter \cite{Blanz1999Morphable} proposed 3D face morphable model (3DMM) to model 3D faces and upon that Dou et al. \cite{Dou2017End} reconstructed 3DMM parameters with a deep neural network. Also, L{\"u}thi et al. \cite{Luthi2018Gaussian} proposed GPMM face model, a generalization of point distribution models.

Different from those model-based methods, GANFIT \cite{Gecer2019GANFIT} and MMFace \cite{Yi2019MMFace} utilize 2D faces to reconstruct 3D faces. 

Gilani and Mian \cite{Gilani2018Learning} and Kim et al. \cite{Kim2017Deep} also proposed data augmentation methods along with their face recognition networks to generate millions of 2D projected images specially for their networks from 3D faces on the ground that their networks use 2D images as the input.

\section{Methodology}

\begin{figure*}[t]
	\centering
	\includegraphics[width=145.60mm]{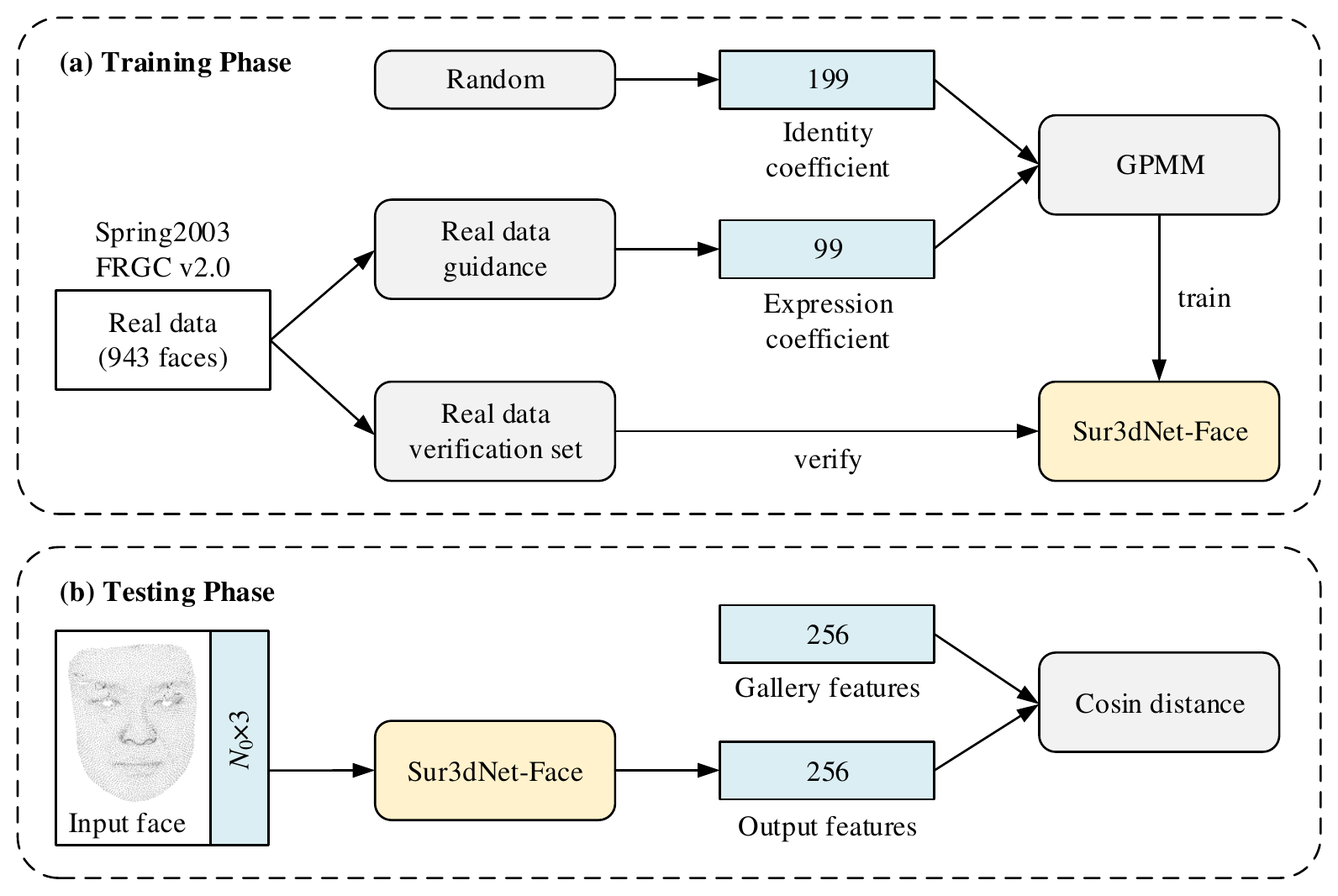}
	\caption{An overview of the proposed few-data guided learning framework for face recognition.}
	\label{fig:flow}
\end{figure*}

In this section, we firstly introduce the architecture of our network, and then the training details are addressed. The overall procedure of the proposed method is shown in Fig.~\ref{fig:flow}.

\subsection{Network Architecture}

Different from the traditional learning-based 3D face recognition methods \cite{Gilani2018Learning, Kim2017Deep}, our method designs an end-to-end network directly inputting the coordinate of point cloud, so as to maintain the advantages of 3D data such as rotation invariance and transformation invariance. The output of our network is a feature vector, and the cosine distance between two feature vectors is calculated to reflect the probability that the two input faces are grabbed from the same subject. 

Specifically, the forward process of our network can be represented as:
\begin{equation}
	f = Sur3dNet\left( \Gamma  \right)
\label{equ:func}
\end{equation}
where $\Gamma  = \left\{ {{x_1},{x_2}, \cdots ,{x_{{N_0}}}} \right\} \in {\mathbb{R}^{{N_0} \times 3}}$ is the unordered input point cloud, ${N_0}$ denotes the number of points, $f \in {\mathbb{R}^{256}}$ is the output features. 

\begin{figure*}[t]
	\centering
	\includegraphics[width=169.89mm]{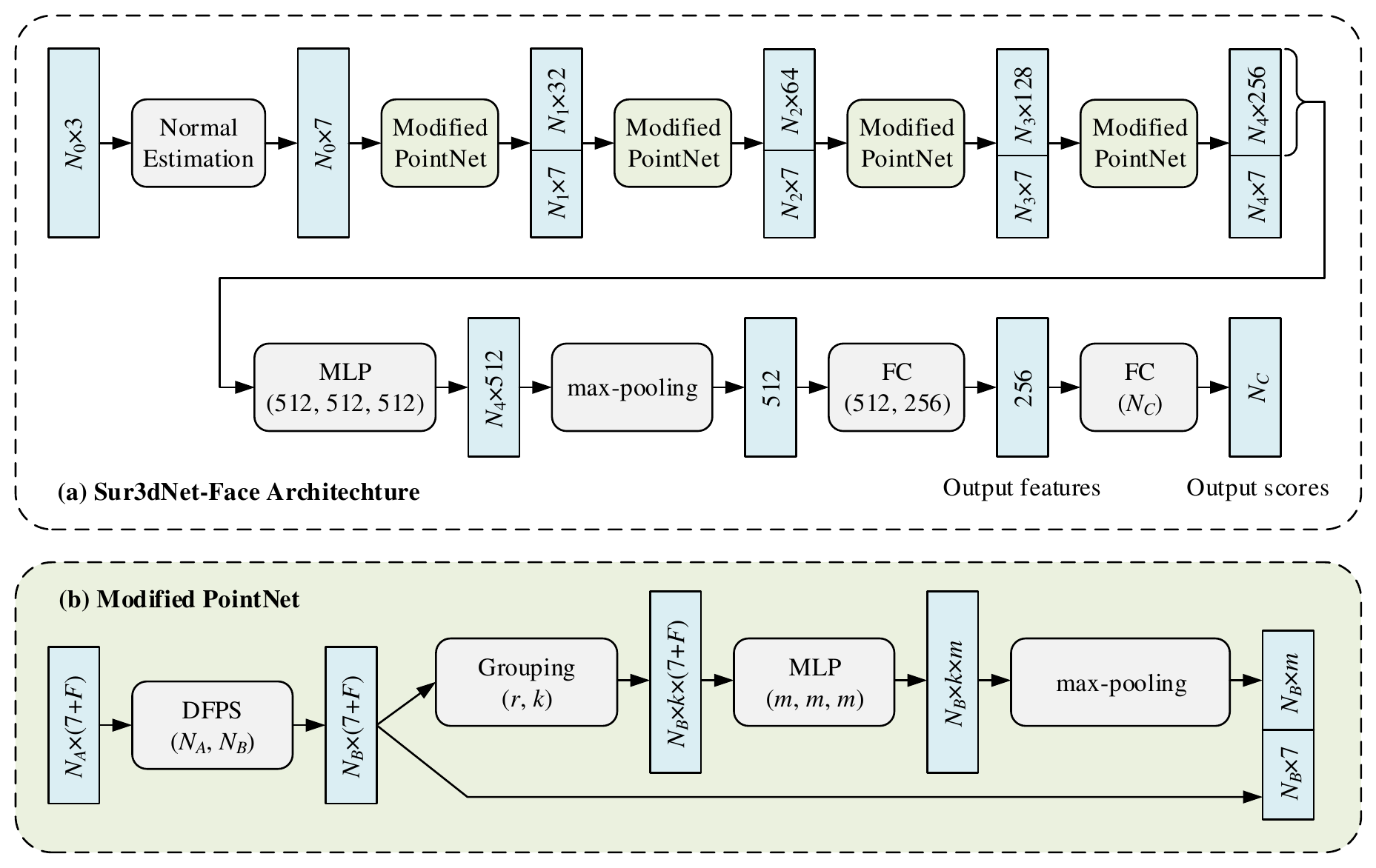}
	\caption{Architecture of our proposed Sur3dNet-Face, an end-to-end point cloud network for 3D face recognition.}
	\label{fig:arch}
\end{figure*}

The architecture of our proposed network is shown in Fig.~\ref{fig:arch} and the submodules in the figure will be introduced in the subsequent subsections.

\subsection{Normal Estimation}

As is known, the normal vector is one of the most important attributes of point clouds. We calculate the normal vectors $n \in {\mathbb{R}^{{N_0} \times 3}}$ of the input point cloud $\Gamma $ through Principal Component Analysis (PCA), during which the corresponding eigen values $e \in {\mathbb{R}^{{N_0} \times 1}}$ can also be derived, which reflect the curvature of the surface. Therefore, the output of the normal estimation submodule is $\left[ {\begin{array}{*{20}{c}}
  \Gamma &n&e 
\end{array}} \right] \in {\mathbb{R}^{{N_0} \times 7}}$.

\subsection{Modified PointNet}

PointNet \cite{Charles2017PointNet} learns a function that maps a set of points to a feature vector, where multi-layer perceptrons (MLPs) are applied to every point individually before a max-pooling layer that aggregates features of all points to a global vector.

The backbone of our architecture is similar to PointNet, as is shown in Fig.~\ref{fig:arch}(b), and the modifications are as follows.

\subsubsection{Ball Query With Physical Size}

The coordinates of all objects are normalized to the range of $\left( { - 1,1} \right)$ in PointNet. However, size is an important attribute of faces, which will be lost in the normalization process. To avoid this, we discard the normalization process and directly input the point cloud with original physical size with the unit of millimeter, and therefore, the radius of ball query in our network should also be measured in millimeters.

\subsubsection{Dithering Farthest Point Sampling (DFPS)}

The traditional farthest point sampling (FPS) algorithm is an iterative process. Given input points $\Gamma  = \left\{ {{x_1},{x_2}, \cdots ,{x_{{N_A}}}} \right\}$, in order to obtain the output set $S = \left\{ {{x_1},{x_2}, \cdots ,{x_{{N_B}}}} \right\}$ with ${N_B}$ points, ${N_B}$ iterations are required, and the formula for each iteration is as follows: 
\begin{equation}
	{x_j} = \arg \max \left( {\min d\left( {{x_i},{x_j}} \right)} \right)
\label{equ:fps}
\end{equation}
where ${x_i} \in S$ is the points already taken out before this iteration, ${x_j} \in \Gamma $ is the point to be taken out in this iteration, $d\left( {{x_i},{x_j}} \right)$ is the Euclidean distance between ${x_i}$ and ${x_j}$, so that ${x_j} \in \Gamma $ is the most distant point relative to the set $S$.

\begin{figure*}
	\centering
	\includegraphics[width=157.09mm]{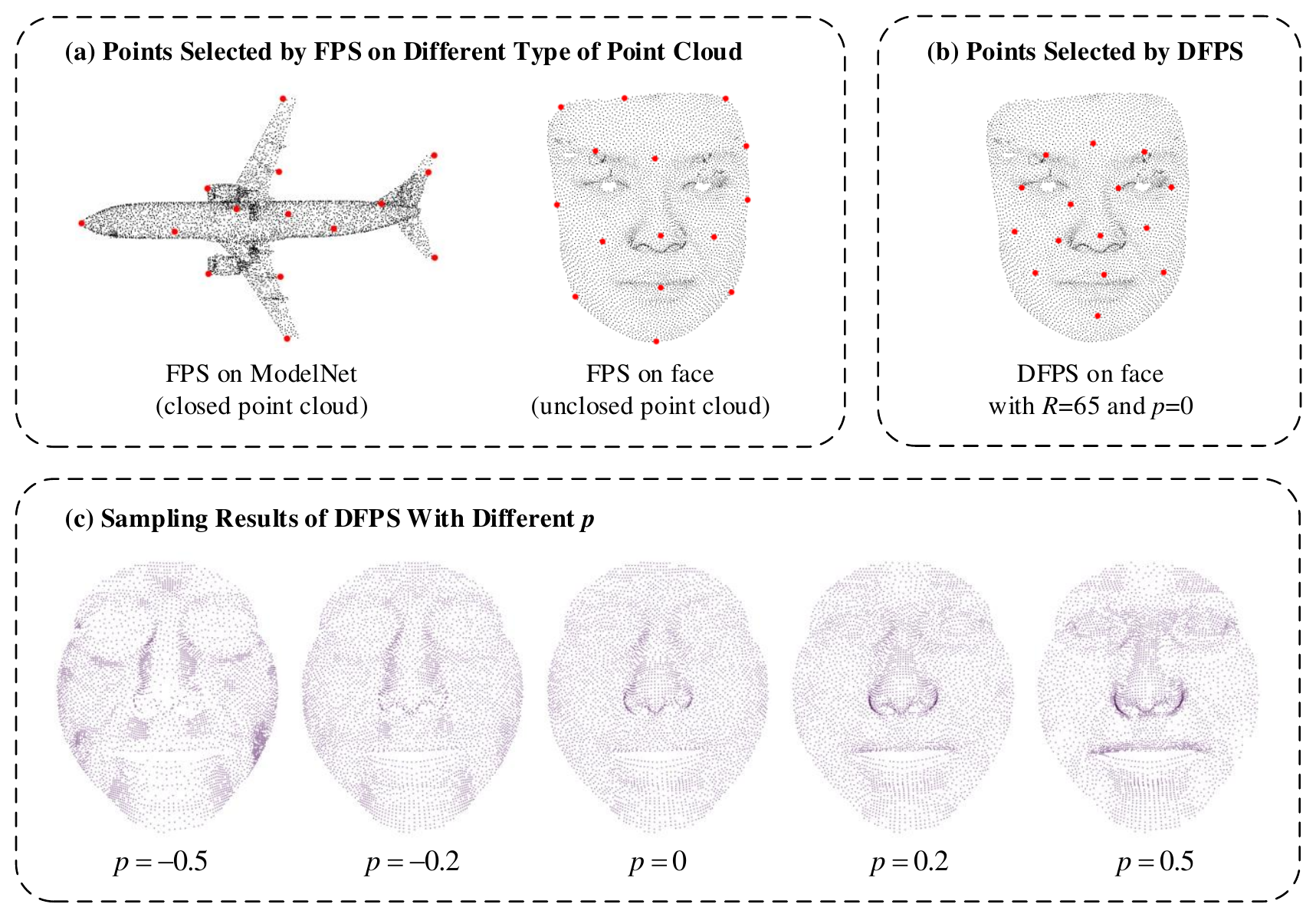}
	\caption{Analysis on the dithering farthest point sampling (DFPS). In (a) and (b), the black points are the input and the first 15 points selected by the algorithm are marked in red. (c) shows the sampling results of DFPS with different $p$, where the corner points are more likely to be selected as $p$ increases.}
	\label{fig:samp}
\end{figure*}

According to our analysis, the main reason for the poor performance of PointNet in face recognition is that FPS preferentially selects points at the edge region of the point cloud. For the closed point clouds generated from CAD model (e.g. ModelNet dataset), this strategy ensures that more corner points can be taken out. However, as compared in Fig.~\ref{fig:samp}(a), in practical scenario, on the ground that the point cloud is grabbed from one direction, barely including one perspective of the object, the edge points are rather unstable. Therefore, the recognition rate of PointNet on the actual measured point cloud is far below expectation, especially on the face with different posture, where the boundary line can be quite different. Meanwhile, for the edge points, the neighbors clustered by ball query are aggregated on one side of the center, leading to high susceptibility of features to pose variation. To solve these problems, we propose dithering farthest point sampling (DFPS) as follows:
\begin{equation}
	{x_j} = \arg \max \left( {\min \lambda d\left( {{x_i},{x_j}} \right)} \right)
\label{equ:dfps}
\end{equation}
where
\begin{equation}
{\lambda _j} = \left\{ \begin{gathered}
  0,d\left( {{x_j},{x_{NT}}} \right) > R \hfill \\
  e_j^p,else \hfill \\ 
\end{gathered}  \right.
\label{equ:lambda}
\end{equation}
where ${x_{NT}}$ is the coordinate of nose tip, ${e_j}$ is the eigen value of ${x_j}$ (as mentioned before in normal estimation subsection), $p$ is the weighting factor, $R$ is the valid radius beyond which the ${\lambda _j}$ will be set to 0 and the corresponding ${x_j}$ will never be selected by DFPS.

DFPS has different behavior in training phase and testing phase. During training, random variations are added to $R$ and $p$ to improve the adaptability of the network. Specifically, in the testing phase $R = 65$ and $p = 0$, while in the training phase, $R$ and $p$ are random numbers in range of $\left( {50,80} \right)$ and $\left( { - 0.2,0.2} \right)$ respectively.

The first row of (\ref{equ:lambda}) ensures that the points output by FPS are mostly selected in the central area that contains abundant facial features, avoiding the influence of unstable edge points, as is shown in Fig.~\ref{fig:samp}(b).

Additionally, it is necessary to explain the second row of (\ref{equ:lambda}) to analyze its principle. As a rule, there is a lot of noise on the real captured point clouds. Owing to the tendency for FPS to select distant points, corner points and noise floating outside the surface of point cloud are prone to be selected. As the eigen value ${e_j}$ reflects the smoothness near the point ${x_j}$, by adjusting the value of $p$, DFPS will tend to select corner points (when $p$ increases) or points on the smooth surface (when $p$ decreases), as is shown in Fig.~\ref{fig:samp}(c). For different measurement system, the smoothness of the output point cloud can be quite different, leading to difficulty in determining $p$, so we adopt a more concise strategy that randomly generates $p$ in training phase. With different $p$, the coordinates of the sampled points will dither slightly, so that the trained network can adapt to different types of noise. Our strategy actually adds more randomness to the traditional FPS, and the experiments show that when supplemented with the proposed dithering strategy, the network gets better results. 

As is shown in Fig.~\ref{fig:arch}(a), there are four modified PointNet layers in our network, and the parameters of each layer are shown in Table~\ref{table:1}.

\begin{table*}
	\renewcommand{\arraystretch}{1.2}
	\setlength{\tabcolsep}{6mm}
	\caption{Parameters of the Four Modified PointNet Layers}
	\centering
    \begin{tabular}{c||c|c|c|c|l}
        \hline
        Parameter & layer 1 & layer 2 & layer 3 & layer 4 & Comments                  \\ \hline
        $N_A$     & 24576   & 4096    & 1024    & 256     & Input point number        \\
        $N_B$     & 4096    & 1024    & 256     & 64      & Output point number       \\
        $r$       & 4       & 8       & 16      & 32      & Radius of ball query (mm) \\
        $k$       & 24      & 32      & 48      & 64      & Ball query point number   \\
        $F$       & 0       & 32      & 64      & 128     & Input feature number      \\
        $m$       & 32      & 64      & 128     & 256     & Output feature number     \\ \hline
    \end{tabular}
	\label{table:1}
\end{table*}

These modifications are seemingly simple, but for the task of face recognition, they are of great significance. In the experiments section, the impacts of these modifications on the recognition rate are compared intuitively through the ablation study. 

\subsection{Other Details}

Our network has no limitation for the number of input points, but in the training phase, point clouds with different sizes cannot be stacked into a batch. In order to use a larger batch size, the input point clouds are firstly downsampled to ${N_0} = 24576$ through random choice. Note that there is no such limitation in the test phase or in practical use, where the ${N_0}$ denotes the actual number of input points. 

Batch normalization layer and dropout layer can significantly improve the performance of the network, so we add both on the FC layers (except the last one), and the dropout rate is 0.5. We use Adam \cite{Kingma2015Adam} optimizer to train the network with the initial learning rate of 1e-3 multiplied by a factor of 0.1 every 10 epochs. Also, the weight decay is set to 1e-4, batch size is 32, and the total number of epochs is 35. 

\subsection{Identification}

Similar to traditional solutions, we take the 256-dimensional vector from the output of the penultimate FC layer as a face representation. After acquiring features of both gallery and probe, we calculate the cosine distance between them, and afterwards, identity of the probe is determined by the gallery with minimum distance. 
 
\subsection{Training Data}

\subsubsection{Gaussian Process Morphable Models}

In order to generate sufficient training data, we use the GPMM face model \cite{Luthi2018Gaussian}, a generalization of point distribution models, which assumes that any shape $\Gamma $ can be represented as a discrete set of points:
\begin{equation}
	\Gamma  = \left\{ {{x_1},{x_2}, \cdots ,{x_N}} \right\}
\label{equ:gpmm1}
\end{equation}
where $N$ denotes the number of points.

Shape $\Gamma $ is represented as a vector $s \in {\mathbb{R}^{3N}}$:
\begin{equation}
	s = {\left[ {\begin{array}{*{20}{c}}
  {{x_{1x}}}&{{x_{1y}}}&{{x_{1z}}}& \cdots &{{x_{Nx}}}&{{x_{Ny}}}&{{x_{Nz}}} 
\end{array}} \right]^T}
\label{equ:gpmm2}
\end{equation}

GPMM model assumes that the shape variation can be modeled using a normal distribution:
\begin{equation}
	s \sim \mathcal{N}(\mu ,\Sigma )
\label{equ:gpmm3}
\end{equation}

According to the theory proposed by \cite{Luthi2018Gaussian}, any face $s \in {\mathbb{R}^{3N}}$ can be expressed as:
\begin{equation}
	s = \bar s + {B_S}\sqrt {{\Lambda _S}} \alpha  + {B_E}\sqrt {{\Lambda _E}} \beta 
\label{equ:deform}
\end{equation}
where the parameter $\alpha $ determines the shape of human face, $\beta $ determines the expression of human face, and $\bar s$ is the mean face model. ${B_S}$ and ${B_E}$ are the basis respectively for shape and expression, and similarly, ${\Lambda _S}$ and ${\Lambda _E}$ are the variance. For $\bar s$, ${B_S}$, ${B_E}$, ${\Lambda _S}$, and ${\Lambda _E}$, we directly adopt the values provided by \cite{Luthi2018Gaussian}.

\subsubsection{GPMM-Based Data Generation}

According to formula (\ref{equ:deform}), the face generated by GPMM depends on two parameters, $\alpha $ and $\beta $, where $\alpha $ determines the identity and $\beta $ determines the expression.

As GPMM is not specially designed for training neural networks, when using faces generated by formula (\ref{equ:deform}) for training, the trained network can recognize the frontal faces. However, in practical application with posture and noise, in order to improve the variety of the data, the formula should be modified as:
\begin{equation}
	s = f\left( {\bar s + {B_S}\sqrt {{\Lambda _S}} \alpha  + {B_E}\sqrt {{\Lambda _E}} \beta  + \delta } \right)
\label{equ:deform2}
\end{equation}
where $f\left( \bullet  \right)$ is the random rotation function, $\delta \sim \mathcal{N}(0,\sigma _\delta ^2)$ is the Gaussian noise, $\alpha \sim \mathcal{N}(0,\sigma _\alpha ^2)$ is the shape coefficient, $\beta \sim \mathcal{N}(0,\sigma _\beta ^2)$ is the expression coefficient. 

\subsubsection{Real Data Guided GPMM-Based Data Generation}

The randomly generating strategy appears difficult to cover the common expressions in the real faces. Taking the disgust expression as an example, we observe that few of the faces with random coefficients look like the disgust. To make our training data further closer to the real data, we propose an enhanced strategy matching the real face with GPMM model, so as to use the real data as a guidance for the expression coefficient to generate training data. 

We denote a pair of real faces in the dataset as $\left( {{\Gamma _N},{\Gamma _E}} \right)$, where ${\Gamma _N}$ and ${\Gamma _E}$ are respectively the neutral face and the expressive face of a certain subject. According to \cite{Gerig2017Morphable}, the registration problem can be solved as:
\begin{equation}
	\begin{gathered}
  {\alpha _N} = \mathop {\arg \max }\limits_\alpha  p\left( \alpha  \right)p\left( {{\Gamma _N}\left| {\alpha ,\bar S} \right.} \right) \hfill \\
  {\beta _E} = \mathop {\arg \max }\limits_\beta  p\left( \beta  \right)p\left( {{\Gamma _E}\left| {\beta ,{\Gamma _N}} \right.} \right) \hfill \\ 
\end{gathered} 
\label{equ:alphabeta}
\end{equation}
where ${\alpha _N}$ is the shape coefficient of both ${\Gamma _N}$ and ${\Gamma _E}$, and ${\beta _E}$ is the expression coefficient of ${\Gamma _E}$.

We calculate ${\beta _E}$ of each expressive face in the Spring2003 subset of FRGC v2.0 in advance. When generating a face, one of ${\beta _E}$ is randomly selected to obtain ${\beta _F}$ as:
\begin{equation}
	{\beta _F} = \lambda {\beta _E} + \left( {1 - \lambda } \right)\beta 
\label{equ:betamix}
\end{equation}
where $\lambda $ is a random value between $\left( {0,1} \right)$, $\beta \sim \mathcal{N}(0,\sigma _\beta ^2)$.

Afterwards, ${\beta _F}$ is substituted into formula (\ref{equ:deform2}) to generate the face, so that the faces generated can cover more expressions similar to real faces in datasets. Experiments show that this strategy can considerably improve the recognition rate of faces with expressions. 

\subsection{Verification Set}

To avoid overfitting, a verification set separated from the training set is usually used in deep learning to verify the performance of the network. In our method, all the training samples are generated from GPMM model upon the same formula, leading to similar distribution. So if we directly take a part of the training samples as the verification set, the loss of the verification set will have almost the same trend as that of the training set, and thus cannot play the role of verifying whether the network is overfitted. Existing public datasets contain too few samples for training, but they are quite sufficient as the verification set. Therefore, we put forward a strategy training the network with the generated data and verifying upon the real data.

Since real data in the verification set do not share the identity with our training data, feature vectors extracted by the network from the real data are used to calculate the cosine distance. Through cosine distance, we can evaluate the recognition performance on the real data verification set through Rank-1 Recognition Rate (RR1), Verification Rate (VR) under ${\text{FAR}} = 1e - 3$, and Area Under ROC Curve (AUC). Taking these three commonly used indicators into account, we define the loss of the verification set as:
\begin{equation}
	loss = 1 - {\text{VR}} \times {\text{RR1}} \times {\text{AUC}}
\label{equ:loss}
\end{equation}

In the experiments, we observe that the loss initially decreases with the training epoch, and then it starts to increase, indicating that the network starts to overfit. The network parameters with the lowest verification loss are saved as the final result of the training process.

\section{Experiments}

In the experiments, the proposed network is implemented on PyTorch \cite{Paszke2019PyTorch} and is training with i7-8700K CPU and two GTX1080TI GPU. 

We use Face Recognition Grand Challenge (FRGC) v2.0 dataset \cite{Phillips2005Overview} and Bosphorus dataset \cite{Savran2008Bosphorus} to evaluate the performance of the proposed face recognition method.

\subsection{Ablation Study}

It takes about 50 hours on our hardware to train the network with the training set of 3000 identities, each with 200 different expressions. In order to make more comparisons in a shorter time, we use training set of 3000 identities for ablation study unless otherwise specified, and the rank-1 recognition rate on Bosphorus dataset excluding occlusion subset and posture subset is the main indicator to evaluate the performance.

\subsubsection{Parameters in DFPS}

\begin{figure*}
    \centering
	\includegraphics[width=176.97mm]{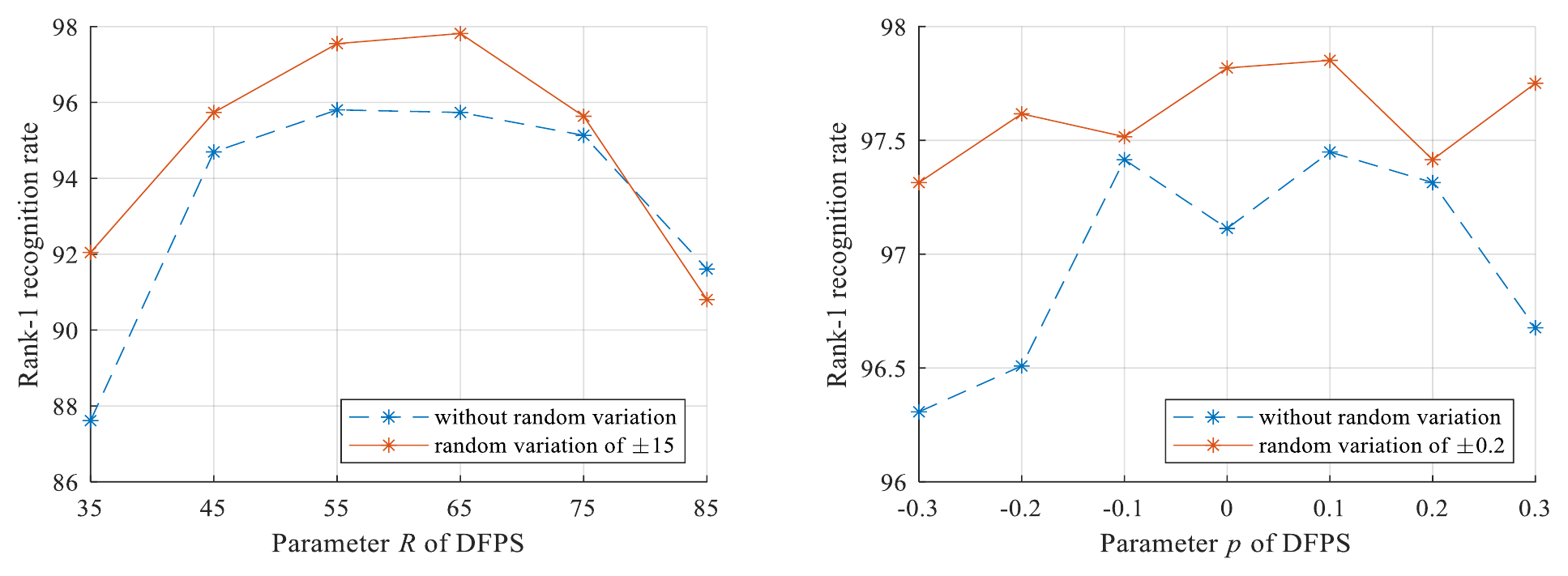}
	\caption{Comparisons on different $R$ and $p$. The results in this figure are obtained without using real data guided generation and verification.}
	\label{fig:asdfps}
\end{figure*}

There are two new parameters introduced in DFPS, namely $R$ and $p$. Fig.~\ref{fig:asdfps} (a) demonstrates the rank-1 recognition rate on Bosphorus dataset by training the network with different $R$ values under $p = 0 \pm 0.2$. It can be seen from the blue line that when $45 \leqslant R \leqslant 75$, the recognition rate has little difference. Meanwhile, when the random $R$ strategy is adopted, where $R$ is added by a random variation between -15 and 15 during training phase, as shown in orange line, the recognition rate is further improved.

Fig.~\ref{fig:asdfps} (b) depicts the rank-1 recognition rate on Bosphorus dataset by training the network with different $p$ values under $R = 65 \pm 15$. In fact, even if we use the same parameters for training, the recognition rate will occasionally show a variation of 0.5\% in each reproduction. Owing to the relatively small impact of parameter $p$ on the results, it appears difficult to determine which is the best value from these experimental results. However, as can be seen from the figure, the random strategy plays a positive role.

Through this ablation study, the parameters of DFPS is determined as follows: in the testing phase $R = 65$ and $p = 0$, while in the training phase, $R$ and $p$ are random numbers in range of $\left( {50,80} \right)$ and $\left( { - 0.2,0.2} \right)$ respectively for each mini-batch.

\subsubsection{Ball Query Radius}

Ball query radius has a significant impact on the recognition rate. We observed that when $r$ and $k$ match a certain proportion, the recognition rate goes higher. Specifically, a smaller $r$ is more suitable for a smaller $k$, and vice versa. Among all the tested combinations, as shown in Table~\ref{table:2}, the combination of $r = 4,8,16,32$ and $k = 24,32,48,64$ gets the best result, and these parameters are also given in Table~\ref{table:1}. Note that the recognition rate in Table~\ref{table:2} is obtained without using real data guided generation and verification.

\begin{table}
	\renewcommand{\arraystretch}{1.2}
	\setlength{\tabcolsep}{6mm}
	\caption{Rank-1 Recognition Rate (RR1) Under Different Combination of $r$ and $k$ on Bosphorus Dataset}
	\centering
    \begin{tabular}{c|c|c}
        \hline
        $r$        & $k$         & RR1     \\ \hline
        3,6,12,24  & 16,24,32,48 & 86.97\% \\
        3,6,12,24  & 24,32,48,64 & 80.53\% \\
        4,8,16,32  & 16,24,32,48 & 95.17\% \\
        4,8,16,32  & 24,32,48,64 & 97.85\% \\
        4,8,16,32  & 40,48,56,64 & 96.66\% \\
        5,10,20,40 & 24,32,48,64 & 94.50\% \\
        5,10,20,40 & 40,48,56,64 & 95.15\% \\ \hline
    \end{tabular}
	\label{table:2}
\end{table}

\subsubsection{Real Data Guided Generation and Verification}

The previous experiments are aimed at network parameters, among which the highest rank-1 recognition rate we achieve is 97.85\%. Afterwards, we make further experiments verifying the effect of training with a small amount of real data. 

There are two real-data-based techniques proposed in this paper, namely, real data guided generation and real data verification set. In the experiments, we randomly select about half of the faces in Spring2003 subset of FRGC v2.0 for real data guided generation and the other half for real data verification set. The results show that our network achieves rank-1 recognition rate of 98.29\% with the former technique, real data guided generation, and when both techniques are applied, the rank-1 recognition rate reaches 98.83\% on Bosphorus dataset.

\subsubsection{Training Data Volume}

\begin{figure}
    \centering
	\includegraphics[width=86.86mm]{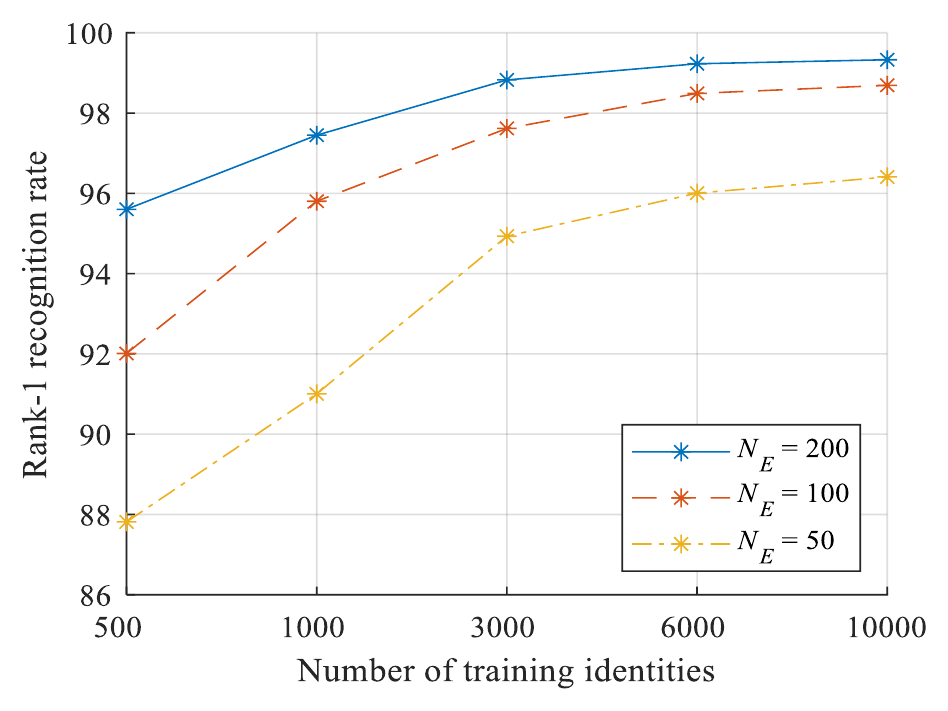}
	\caption{Comparisons on different volume of training data. ${N_E}$ denotes the number of generated expressions for each identity.}
	\label{fig:datavol}
\end{figure}

The training data can be arbitrarily generated, but due to the limitation of hardware performance, we test 10000 identities at most, each containing 200 expressions, and the recognition rate on Bosphorus dataset is displayed in Fig.~\ref{fig:datavol}.

From the experimental results, the rank-1 recognition rate reaches 99.33\% of training with 10000 identities each with 200 expressions. Also, the increasing trend of recognition rate in the figure indicates that if a larger data volume is used, a slightly higher recognition rate can possibly be obtained. 

\subsection{Comparisons With Other Methods}

\subsubsection{Results on FRGC v2.0}

FRGC v2.0 dataset \cite{Phillips2005Overview} containing 4007 scans of 466 subjects in total is divided into three partitions as Spring2003 subset, Fall2003 subset, and Spring2004 subset. Protocol \cite{Phillips2005Overview} uses the Spring2003 for training and the remaining for testing. We follow this protocol taking Spring2003 as real data to train the network. Specifically, we randomly select about half of the faces in Spring2003 for real data guidance and the other half for real data verification set.

The comparisons between our proposed method and the state-of-the-art methods on FRGC v2.0 dataset are shown in Table~\ref{table:3}, where the most representative face recognition methods are compared. Some methods use the corresponding 2D photos of the 3D faces, which we denote as (2D+3D). Also, there are some methods using fine-tuning to further improve the recognition rate, which are marked with (FT).

It can be seen from the table that some deep-learning-based methods have achieved satisfactory recognition rate in recent years, especially those with fine-tuning that can considerably improve the recognition rate. However, in the actual application scenario, due to the dynamic changes of the face gallery, the effect of fine-tuning is much lower than expectation. Among those methods without using fine-tuning, our method is quite competitive, which achieves rank-1 recognition rate of 98.85\% and verification rate of 96.75\% on FRGC v2.0 dataset.

As is known, whether deep learning can achieve good results, to a great extent, depends on training data. Different from those methods training with more and more data to get a slightly higher recognition rate, the only real data we use in the training process are the 943 faces in Spring2003 subset. With such a small amount of data, we achieve a higher recognition rate than some of the latest methods, which proves the effectiveness and the potentiality of our method.

\begin{table}
	\renewcommand{\arraystretch}{1.2}
	\setlength{\tabcolsep}{4mm}
	\caption{Rank-1 Recognition Rate (RR1) and Verification Rate (VR) Under ${\text{FAR}} = 1e - 3$ on FRGC v2.0 Dataset}
	\centering
    \begin{tabular}{lcc}
        \hline
        Method                                                    & RR1     & VR      \\ \hline
        Mian et al. \cite{Mian2008Keypoint} (2008) (2D+3D)        & 96.10\% & 98.60\% \\
        Huang et al. \cite{Huang20123D} (2012)                    & 97.60\% & 98.40\% \\
        Liu et al. \cite{Liu2013Learning} (2013)                  & 96.94\% & 90.00\% \\
        Elaiwat et al. \cite{Elaiwat2015Curvelet} (2015)          & 97.10\% & 99.20\% \\
        Lei et al. \cite{Lei2016Two} (2016)                       & 96.30\% & 98.30\% \\
        Al-Osaimi \cite{Osaimi2016Novel} (2016)                   & 96.49\% & 98.69\% \\
        Ouamane et al. \cite{Ouamane2017Efficient} (2017)         & -       & 96.65\% \\
        Ouamane et al. \cite{Ouamane2017Efficient} (2017) (2D+3D) & -       & 98.32\% \\
        Gilani et al. \cite{Gilani2018Dense} (2018)               & 98.50\% & 98.70\% \\
        Gilani and Mian \cite{Gilani2018Learning} (2018)          & 97.06\% & -       \\
        Gilani and Mian \cite{Gilani2018Learning} (2018) (FT)     & 99.88\% & -       \\
        Cai et al. \cite{Cai2019Fast} (2019) (FT)                 & 100\%   & 100\%   \\ \hline
        Ours                                                      & 98.85\% & 96.75\% \\ \hline
    \end{tabular}
	\label{table:3}
\end{table}

\subsubsection{Results on Bosphorus}

Bosphorus dataset \cite{Savran2008Bosphorus} includes totally 4666 scans collected from 105 subjects (60 men and 45 woman aged between 25 and 35) with poses changes, expression variations, and typical occlusions. The yaw rotation of faces is from 10 degrees to 90 degrees in Bosphorus dataset. This paper does not involve occlusion and large posture, so the occlusion subset and the posture subset are excluded. The neutral scans with file name containing N\_N\_0 are used to form the gallery features.

The comparisons between our method and the state-of-the-art methods on Bosphorus dataset are shown in Table~\ref{table:4}. In order to make a fair comparison, the recognition rate under the same subsets of Bosphorus is displayed. For the methods that do not provide the result of each subset, the recognition rate of the complete dataset is shown.

\begin{table}
	\renewcommand{\arraystretch}{1.2}
	\setlength{\tabcolsep}{4mm}
	\caption{Rank-1 Recognition Rate (RR1) and Verification Rate (VR) Under ${\text{FAR}} = 1e - 3$ on Bosphorus Dataset}
	\centering
    \begin{tabular}{lcc}
        \hline
        Method                                                    & RR1     & VR      \\ \hline
        Mian et al. \cite{Mian2008Keypoint} (2008) (2D+3D)        & 96.40\% & -       \\
        Huang et al. \cite{Huang20123D} (2012)                    & 97.00\% & -       \\
        Liu et al. \cite{Liu2013Learning} (2013)                  & 95.63\% & 81.40\% \\
        Berretti et al. \cite{Berretti2013Matching} (2013)        & 95.67\% & -       \\
        Elaiwat et al. \cite{Elaiwat2015Curvelet} (2015)          & -       & 91.10\% \\
        Lei et al. \cite{Lei2016Two} (2016)                       & 98.90\% & -       \\
        Al-Osaimi \cite{Osaimi2016Novel} (2016)                   & 92.41\% & 93.5\%  \\
        Ouamane et al. \cite{Ouamane2017Efficient} (2017) (2D+3D) & -       & 96.17\% \\
        Gilani et al. \cite{Gilani2018Dense} (2018)               & 98.5\%  & -       \\
        Gilani and Mian \cite{Gilani2018Learning} (2018)          & 96.18\% & -       \\
        Gilani and Mian \cite{Gilani2018Learning} (2018) (FT)     & 100\%   & -       \\
        Cai et al. \cite{Cai2019Fast} (2019) (FT)                 & 99.75\% & 98.39\% \\ \hline
        Ours                                                      & 99.33\% & 97.70\% \\ \hline
    \end{tabular}
	\label{table:4}
\end{table}

\section{Discussions}

Although some existing methods have achieved satisfactory recognition rate, our method has some notable advantages, upon that we make further discussions.

\subsection{Computational Complexity}

Our method designs an end-to-end network directly inputting the point clouds, which is faster than those methods requiring a complex preprocessing. For example, \cite{Lei2016Two} reports that they need 3.16\,s for preprocessing, including detection and alignment, region segmentation, model registration, and other steps; and \cite{Kim2017Deep} reports that they need 6.08\,s to generate the 2D images including depth map, azimuth map, and elevation map from 3D point clouds.

On the hardware platform described in experiments section, in the training phase with the faces generated in advance, the forward time and backward time of our network for each mini-batch (with batch size of 32) is 0.208\,s and 0.095\,s, respectively. When being applied to actual applications, it takes about 0.035\,s to detect the nose tip and calculate the normal vectors, and 0.105\,s to extract the feature of a single face using a single GPU, and there is no additional preprocessing required. 

\subsection{Data Generation and Overfitting}

There is a common problem in existing deep-learning-based face recognition methods, that is, owing to the lack of data, more faces need to be generated by interpolating existing datasets with each other \cite{Gilani2018Learning, Kim2017Deep}. Although the generated faces used for training are considered different, they have implicit overlaps with the test set, meaning that the results may be obtained to some extent by overfitting.

Comparably, the training data in our method are absolutely generated from GPMM model, which have little intersection with real datasets. Therefore, what we listed are actually cross-dataset results without any fine-tuning, which are closer to the results in actual application than results of those methods that take part of the datasets to generate training data and use the other part to test.

\section{Conclusion}

An end-to-end deep learning network entitled Sur3dNet-Face for point-cloud-based 3D face recognition is presented in this paper, along with the concrete approach for training. Coupled with real data guided generation and real data verification set, a few-data guided learning framework based on Gaussian process morphable model is proposed, upon that the common problem in 3D face deep learning of lacking training data is overcome.

Different from existing methods training with more and more data to get a slightly higher recognition rate, the only real data we use in the training process are the 943 faces in Spring2003 subset of FRGC v2.0. With such a small amount of data, we achieve a higher recognition rate than some of the latest methods, which proves the effectiveness and the potentiality of our method.

Furthermore, the ablation study of our method have been analyzed, and the validity has been proved by experiments. Without any fine-tuning on test set, the Rank-1 Recognition Rate (RR1) and Verification Rate (VR) are achieved as follows: 98.85\% (RR1) and 96.75\% (VR) on FRGC v2.0 dataset, and 99.33\% (RR1) and 97.70\% (VR) on Bosphorus dataset.

In the future research, we will optimize the process of data generation to generate more occluded and angled faces, so as to further improve the applicability of our method.


%





\ifCLASSOPTIONcaptionsoff
  \newpage
\fi



\bibliographystyle{IEEEtran}
\bibliography{egbib}
\end{document}